\documentclass[letterpaper]{article} 
\usepackage[submission]{aaai2026}  
\usepackage{times}  
\usepackage{helvet}  
\usepackage{courier}  
\usepackage[hyphens]{url}  
\usepackage{graphicx} 
\usepackage{natbib}  
\usepackage{caption} 
\usepackage{enumitem}
\usepackage{graphicx}  
\usepackage{caption}   
\usepackage{algorithm}
\usepackage{algorithmic}
\usepackage{adjustbox}
\usepackage{newfloat}
\usepackage{listings}
\DeclareCaptionStyle{ruled}{labelfont=normalfont,labelsep=colon,strut=off} 
\floatstyle{ruled}
\newfloat{listing}{tb}{lst}{}
\floatname{listing}{Listing}
\title{Positional Bias in Multimodal Embedding Models: Do They Favor the Beginning, the Middle, or the End?}

\author {
    Kebin Wu\textsuperscript{\rm 1},
    Fatima Albreiki\textsuperscript{\rm 1}
}
\affiliations {
    \textsuperscript{\rm 1}Technology Innovation Institute (TII), Abu Dhabi, UAE \\
    kebin.wu@tii.ae, fatima.albreiki@tii.ae
}

\usepackage{bibentry}

\makeatletter
\def\showauthors@on{T}
\makeatother

\newcommand{\etal}{\emph{et al.}\ }

\begin{document}

\maketitle

\begin{abstract}
Positional bias—where models overemphasize certain positions regardless of content—has been shown to negatively impact model performance across various tasks. While recent research has extensively examined positional bias in text generation models, its presence and effects in representation models remain underexplored. Even less is known about such biases in multimodal models. In this work, we investigate positional bias in multimodal representation models, specifically in the context of image-text retrieval. We begin by distinguishing between context importance and positional bias, and then assess the presence and extent of positional bias across different models and datasets. Our experiments demonstrate that positional bias is prevalent in multimodal models, but manifests differently across modalities: text encoders tend to exhibit bias toward the beginning of the input, whereas image encoders show bias at both the beginning and end. Furthermore, we find that this bias arises from, or is amplified by, a combination of factors, including the positional encoding scheme, training loss, context importance, and the nature of using image-text pairs in multimodal training.
\end{abstract}

\begin{links}
    \link{Code}{https://github.com/tiiuae/PosBias/}
\end{links}


\section{Introduction}

Transformer-based models have achieved notable success, especially in natural language processing, leading to the development of numerous language models. However, emerging research indicates that their ability to capture contextual information is influenced by the position of that information within the input sequence—an issue known as positional bias. For example, Liu \etal~\cite{liu2024lost} showed that models often prioritize content at the beginning or end, while neglecting the middle—a phenomenon termed “lost in the middle.” This weakens their reasoning abilities and introduces instability in evaluations. In response, many studies have explored the underlying causes of positional bias and proposed mitigation strategies.

\begin{figure*}[tbp!]
    \centering
    \includegraphics[width=0.99\textwidth]{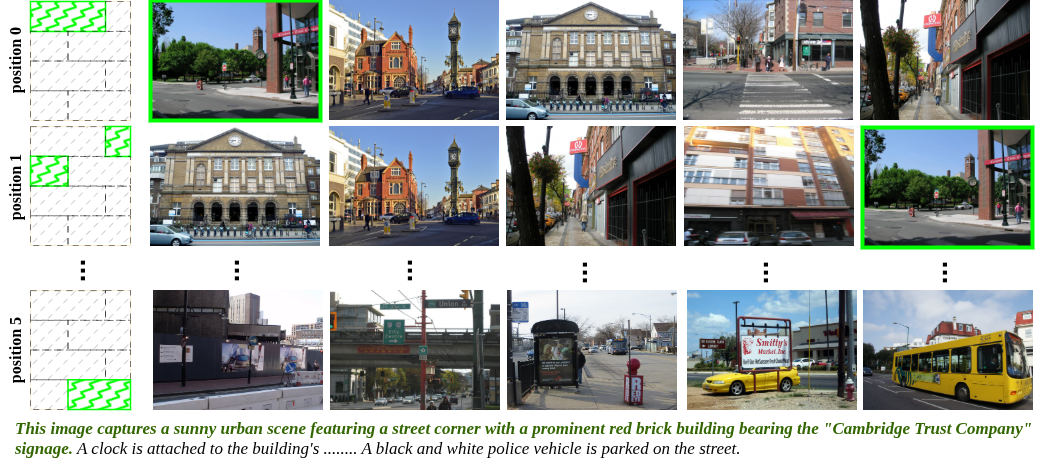}
    \caption{Top-5 text-to-image retrieval results. The caption is divided into six segments. Each row corresponds to a different position where only Segment 0 (highlighted in dark green) is placed, while the remaining positions are masked. Across the six rows, Segment 0 is shifted through all six possible positions.}
    \label{fig:intro}
\end{figure*}

Despite recent progress, research on positional bias has largely focused on text generation. We extend this by analyzing positional bias in multimodal representation learning, specifically within cross-modal retrieval tasks. Our study examines representative multimodal embedding models such as CLIP~\cite{radford2021learning} and evaluates positional bias in their text and image encoders separately. Figure~\ref{fig:intro} shows a text-to-image retrieval example where shifting the position of the ground-truth text within the query significantly alters the retrieved results.

We begin by distinguishing between positional bias and contextual importance. To investigate positional bias, we segment the image or text into multiple parts and iteratively shift a selected segment across different positions, while replacing the remaining segments with either dummy perturbations or modality- and model-specific masking content. This approach reveals substantial positional biases in both the image and text encoders of multimodal models. We extend the analysis to several multimodal models and investigate the prevalence and causes of such bias in multimodal representation learning.


In this paper, we make the following contributions. First, we reveal the presence of positional bias in multimodal models—a phenomenon not previously reported, to our knowledge. Second, we empirically characterize distinct bias patterns: in text encoders, performance is consistently higher when a segment appears at the beginning of the sequence; in image encoders, bias is more variable, favoring either the beginning or both ends. Third, through extensive experiments, we show that this bias persists across a wide range of models, regardless of data distribution, positional encoding, training length, model size, resolution, patch size, loss function, or vision encoder architecture. Finally, we identify several contributing factors that may drive or modulate positional bias.

\section{Related Work}

\subsection{Positional Bias in Text Generation}
Positional bias has been extensively studied in text generation models, especially in the context of long-context language modeling. Liu \etal~\cite{liu2024lost} identified the ``lost in the middle'' phenomenon, where language models underperform when relevant information appears mid-sequence. Hsieh \etal~\cite{hsieh2024found} found that large language models often exhibit a U-shaped attention pattern, emphasizing the beginning and end of sequences over the middle, regardless of semantic importance. Several mitigation strategies have been proposed, including attention calibration~\cite{hsieh2024found}, data augmentation~\cite{he2023never}, differential attention mechanisms~\cite{ye2024differential}, and replacing causal attention with bidirectional variants~\cite{wang2024eliminating}. While some studies have extended this analysis to vision-language models~\cite{tian2025identifying}, their focus remains on text generation tasks like image captioning.

\subsection{Positional Bias in Text Representation}
Positional bias has been observed not only in generation tasks but also in text representation models used for retrieval and classification. Coelho \etal~\cite{coelho2024dwell} investigated this bias in web document retrieval and found that such models tend to prioritize the beginning of documents. They attribute this effect to contrastive training and the inverted pyramid structure commonly found in writing. Similar head-biased tendencies were found in Named Entity Recognition (NER), Part-of-Speech (POS) tagging~\cite{ben2024impact}, and extractive question answering~\cite{ko2020look}. Goel  \etal~\cite{goel2024quantifying} analyzed this bias through semantic similarity measurement, showing consistent overemphasis on early tokens across diverse embedding models. These biases have been hypothesized to result from characteristics of the training data and preprocessing steps such as input truncation. In this paper, we extend the investigation of positional bias to the multimodal setting by empirically examining its existence, patterns, and the underlying causes or contributing factors.

\subsection{Multimodal Representation Learning}
Multimodal representation learning models, particularly CLIP~\cite{radford2021learning} and its variants, align visual and textual inputs in a shared embedding space using contrastive objectives. While CLIP demonstrates strong performance, it relies on large batch sizes and is limited to short textual inputs. SigLIP~\cite{zhai2023sigmoid} mitigates the batch size issue by replacing the contrastive loss with a sigmoid-based alternative. To address text length limitations, Long-CLIP \cite{zhang2024long} introduces a knowledge-preserving positional embedding stretching strategy, extending the token limit to 248 and fine-tuning the model on a dataset specifically curated for long image-text pairs \cite{chen2024sharegpt4v}. In contrast to such approaches that rely on absolute positional encoding, TULIP \cite{najdenkoska2024tulip} adopts relative positional encoding, enabling the model to process text of arbitrary length. This design allows CLIP-like models to make full use of longer captions, thereby enhancing their ability to capture fine-grained and detailed semantic information embedded in extended textual contexts.

Other multimodal models, such as LLaVA \cite{liu2023visual} and BLIP-2 \cite{li2023blip}, have gained prominence for enabling complex, instruction-based multimodal generation tasks. However, in this work, we focus on CLIP and its variants, which are designed for multimodal representation learning through contrastive alignment of image and text embeddings.  This focus is motivated by the fact that CLIP-style models frequently serve as the key backbone for more advanced multimodal systems, including those used in generative settings. In addition, cross-modal retrieval plays a crucial role in multimodal generation pipelines that incorporate Retrieval-Augmented Generation (RAG) techniques \cite{wu2024visual, xia2024mmed}. As such, a deeper understanding of CLIP-like models in the context of cross-modal retrieval is critical for advancing the broader landscape of vision-language learning. 

A more detailed discussion of related work is provided in the supplementary material.
\begin{figure*}[tbp!]
    \centering
    \includegraphics[width=0.99\textwidth]{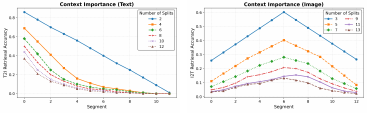}
    \caption{Contextual importance analysis across modalities (image and text). The x-axis shows positions, and colors indicate different number of splits.} 
    \label{fig:context_importance}
\end{figure*}

\section{Methods}

\subsection{Context Importance}\label{sec:context-importance}
Positional bias is sometimes confounded with contextual importance, as models may attend more to certain positions not solely due to their placement, but because those positions often contain semantically important content. To disentangle these factors, we first identify the regions of contextual importance for the text and image encoders separately.

To evaluate contextual importance in the \textbf{text encoder} of a trained CLIP model or its variants, we divide the input tokens into uniform segments. For each trial, we retain only the tokens within one segment and mask the rest using the model’s padding token. The masked input is then passed through the text encoder to obtain its representation, with the image input held constant to avoid cross-modal interference. A similar approach is used for the \textbf{image encoder}: one segment of the image is kept while the others are masked using CLIP’s RGB mean values ([0.481, 0.458, 0.408]), while the text input remains fixed to isolate the contribution of each image segment.

Figure~\ref{fig:context_importance} demonstrates the results for Long-CLIP (ViT-B/16) on the Urban1k dataset~\cite{zhang2024long} under various segment split configurations. Since different numbers of splits result in different segment granularities, we apply interpolation to align the segment-wise importance values onto a common scale for visually coherent comparison. For the contextual importance of text (left), retrieval accuracy is highest with the first segment and decreases monotonically with later ones, reflecting the common writing pattern where key information appears early. For the image modality (right), the central region shows the highest importance, with a gradual drop toward the edges—consistent with the typical compositional bias in natural images, where subjects are centered and the top/bottom often contain less informative background like sky or ground. These trends remain consistent across segment granularities, and generalize across datasets and model architectures (see Section~\ref{sec:supp_context_importance} in the supplementary material). Building on these observations, we next investigate positional bias, with a specific focus on whether regions of bias and contextual importance overlap.

\vspace{0.1ex}
\subsection{Positional Bias}\label{sec:method-bias}
To investigate positional bias in the \textbf{text encoder} of a multimodal model, we adopt two strategies: text perturbation and token masking. In the perturbation approach, the input text is divided into sub-texts, one of which is moved across positions while the others are replaced with Lorem Ipsum dummy placeholders to minimize semantic interference \cite{goel2024quantifying}. To further reduce semantic interference, we apply token masking: the tokenized input is split into segments, and one segment is shifted across positions while the rest are masked with the model’s padding token. In both cases, the image input remains fixed, and we evaluate retrieval accuracy for each manipulated position. Similarly, to assess positional bias in the \textbf{image encoder}, a single visual segment is isolated and moved across spatial locations, with all other regions masked using CLIP’s RGB mean values (see Figure~\ref{fig:image-to-text} in the supplementary material).

Unlike previous studies on positional bias—where the order of multiple documents is shuffled while preserving the internal content of each—we isolate a single segment of the input sequence and systematically move it across different positions. This design is motivated by two key considerations. First, the models examined in our study operate within a fixed and limited context window; including the full input sequence would occupy all available positions, leaving no room to vary token placement for positional analysis. Second, prior work typically concatenates several candidate inputs into a single sequence, where only one is relevant to the query and the others function as distractors. In contrast, in our setting, all image or text segments are closely aligned with the query, making it inappropriate to treat any segment as an irrelevant perturbation. Consequently, simple shuffling would not isolate positional effects in a meaningful way. Importantly, using a single segment per input does not compromise generalization, as our goal is not to measure absolute retrieval accuracy, but rather to analyze how retrieval performance varies with positional changes.

\begin{figure*}
    \centering
    \includegraphics[width=0.98\textwidth]{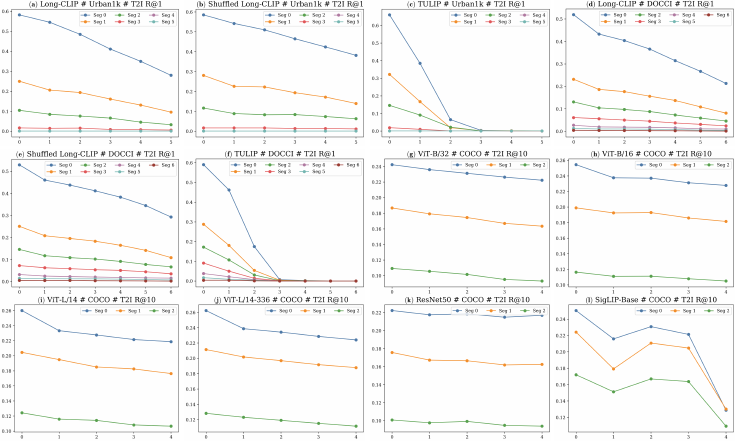}
    \caption{Positional bias analysis of text encoders across multimodal models and datasets. Each figure title follows the format: model name \# dataset \# metric. The x-axis shows positions, and colors indicate different segments.}
    \label{fig:bias-text}
    \vspace{-10pt}
\end{figure*}

\section{Experiments}

\subsection{Experimental Setting}
{\bf Dataset. }We use three image-text retrieval datasets: Urban1K \cite{zhang2024long} and DOCCI \cite{onoe2024docci} for long captions, and COCO \cite{lin2014microsoft} for short captions. Unlike prior work focusing only on long texts, we analyze positional bias across both short and long inputs.

\noindent{\bf Models.} In this work, we empirically assess positional bias across models trained under diverse settings. This enables us to evaluate the (1) generality of the observed bias and to (2) provide evidence that either supports or challenges prevailing assumptions on the causes of positional bias, such as position encoding schemes, loss functions, training stages, data distribution, and preprocessing methods. Our analysis includes the following models: \\

\begin{itemize}[leftmargin=*, itemsep=0pt, topsep=0pt, parsep=0pt]
    \item \textbf{Long-CLIP (ViT-B/16)}~\cite{zhang2024long}: Fine-tuned on a long-caption dataset~\cite{chen2024sharegpt4v} with a 248-token context window and absolute positional encoding.
    \item \textbf{TULIP (ViT-L/14)}~\cite{najdenkoska2024tulip}: Uses relative positional encoding for long captions, allowing comparison with Long-CLIP’s absolute encoding.
    \item \textbf{Shuffled Long-CLIP (ViT-B/16)}: To examine the role of data structure in positional bias, we train a variant of Long-CLIP on shuffled captions. Each original caption is split into sub-captions at sentence boundaries (e.g., “.”, “?”, “!”). The sub-captions within the same caption are then randomly reordered (with internal word order preserved), concatenated, and used for training. All other settings are kept identical to Long-CLIP.
    \item \textbf{CLIP}~\cite{radford2021learning}: Official models (ViT-B/32, B/16, L/14, L/14-336, ResNet-50) are included to assess bias in short-caption settings and to examine the effects of architecture, patch size, resolution, and model size.
    \item \textbf{SigLIP-Base}~\cite{zhai2023sigmoid}: This model differs from CLIP primarily in its loss function and is included to assess the role of contrastive loss in inducing positional bias~\cite{coelho2024dwell}.
\end{itemize}
\vspace{1ex}

\noindent We use official checkpoints for TULIP, CLIP, and SigLIP. Long-CLIP is reproduced using the official recipe with two changes: mixed precision is disabled, and training is run on 4 GPUs with batch size 128. Shuffled Long-CLIP uses the same setup with shuffled sub-captions.

\noindent{\bf Evaluation Metric.} Most experiments in this paper focus on cross-modal retrieval tasks. For the long-caption dataset, we report Recall@1 for both image-to-text (I2T) and text-to-image (T2I) retrieval. For the short-caption dataset COCO, we instead report Recall@10, as Recall@1 is generally too low to support meaningful comparisons. However, T2I (respectively, I2T) is considered when evaluating positional bias in the text (respectively, image) encoder, as the bias analysis involves modifying the text (image) input while keeping the other modality unchanged. In investigating the cause of positional bias, we also conducted experiments on a classification task, where top-1 classification accuracy is utilized as the evaluation metric.

\subsection{Experimental Results on Positional Bias}\label{sec:bias-main}
In this section, we first examine positional bias in the multimodal model Long-CLIP (ViT-B/16) on the long-caption dataset Urban1K. We extend the analysis to other datasets and models in a later section, as part of our discussion on the reasons of positional bias.

\noindent{\bf Textual Positional Bias.}
On Urban1K, we divide each caption into six segments, based on data analysis indicating that six is the modal number of sub-captions. In Figure~\ref{fig:bias-text}(a), we present results using a token masking strategy, where each color represents the specific segment (among the six) that is selected and moved across different positions in the input sequence to evaluate positional effects. Here, the step size matches the segment length, resulting in six valid positions. We observe that retrieval accuracy is consistently highest when any given segment is placed at the beginning of the sequence, regardless of its original position in the caption. This indicates a strong positional bias favoring the beginning. When applying the text perturbation method, we observe a similar trend, with the strongest bias again appearing at the beginning of the sequence (see Figure~\ref{fig:dummy-token-comparsion} in the supplementary material). Therefore, for brevity, we include only the token masking results in the main paper and provide perturbation-based results in Figure~\ref{fig:bias-text-perturbation} in the supplementary material. In addition to positional effects, Figure~\ref{fig:bias-text}(a) also reveals context importance. By comparing different segments placed at the same fixed position (i.e., comparing different colors along a vertical slice), we find that the first segment consistently yields the highest retrieval performance. 
 
\begin{figure*}[t]
    \centering
    \includegraphics[width=0.98\textwidth]{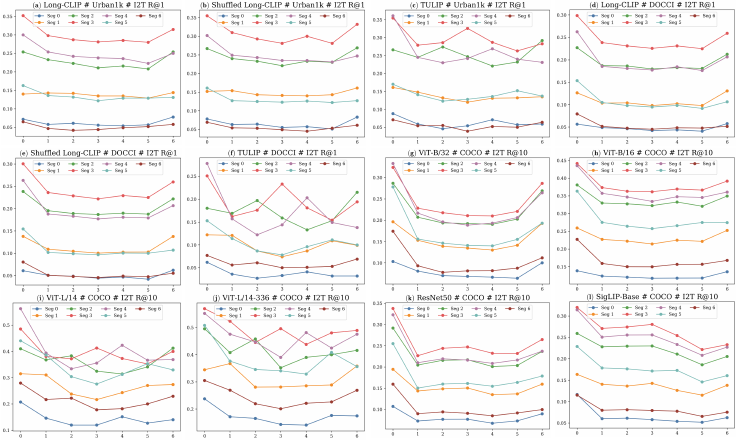}
    \caption{Position bias analysis of image encoders across models and datasets. Each figure title follows the format: model name \# dataset \# metric. The x-axis shows positions, and colors indicate different segments.}
    \label{fig:bias-image}
\end{figure*}

\noindent{\bf Visual Positional Bias.} Extending the methodology used to study positional bias in text encoder, we examine positional effects on the image side by setting the split number to 7. Choosing an odd number allows the central segment, which often contains the most informative visual content as shown in Section~\ref{sec:context-importance}, to occupy a distinct middle position and enables the detection of potential central bias. Figure~\ref{fig:bias-image}(a) shows the positional bias in the image encoder of the Long-CLIP (ViT-B/16) model on Urban1K. The results reveal a clear bias toward both the beginning and end positions in most segments (observed in 5 out of 7 cases), with the bias at the beginning being notably stronger. Given that the central segment is the most semantically informative for images as discussed previously, the tendency to favor the beginning and end, particularly the beginning, is unexpected.

\subsection{Investigating Reasons for Positional Bias}

Previous work has attributed positional bias—mainly in text generation and representation models—to factors such as human writing style \cite{coelho2024dwell}, training objectives \cite{coelho2024dwell}, data preprocessing \cite{goel2024quantifying}, positional embeddings, and causal attention \cite{wang2024eliminating}. Here, we conduct experiments to evaluate these assumptions and either support or challenge them in the context of \textbf{multimodal} models.

\vspace{0.3ex}
\noindent{\bf Data Distribution.}  
It is commonly assumed that positional bias in representation models arises from the data distribution: when certain image regions or text segments consistently enhance performance, models learn to favor those positions, thus developing positional bias towards them. However, based on our findings on context importance (Section~\ref{sec:supp_context_importance}) and positional bias (Section~\ref{sec:bias-main}), we argue that regions with high contextual importance do not always overlap with those exhibiting positional bias. 

For images, central regions are the most semantically informative, yet the observed bias favors the beginning and the end. \textit{This misalignment does not support the initial hypothesis}. On the text side, moving any segment to the beginning consistently improves retrieval accuracy over its original position. Since the most informative segment also tends to appear at the beginning, aligning with the strongest positional bias, it remains unclear whether importance alone drives the bias. To disentangle these factors, we introduce Shuffled Long-CLIP (ViT-B/16), which randomly reorders sub-captions for training to reduce position-specific cues on the text side. Yet, the model still shows a strong bias toward the beginning, as shown in Figure~\ref{fig:bias-text}(b). However, compared to the original Long-CLIP, Shuffled Long-CLIP shows a less pronounced drop in accuracy across positions. In particular, shifting the first segment from the beginning to the end causes a drop of 0.199 (from 0.581 to 0.382) in Shuffled Long-CLIP, whereas the original Long-CLIP sees a larger drop of 0.303 (from 0.581 to 0.278). \textit{These findings suggest that contextual importance contributes to positional bias in text representations to some extent, but it does not fully account for the observed bias.}

Notably, these findings are not limited to a single dataset. Similar patterns of positional bias (Figure~\ref{fig:bias-text}(d) for text and Figure~\ref{fig:bias-image}(d) for image) and contextual importance (Figure~\ref{fig:context_importance_more} in the supplementary material) are observed on dataset DOCCI, which features more diverse images and human-annotated captions than Urban1K. For DOCCI, we set the number of text splits to seven, matching the average number of sub-sentences per caption. 

\vspace{0.3ex}
\noindent{\bf Positional Encoding.} 
Encoding positional information is essential for contextual understanding. CLIP and its variants typically achieve this using learnable absolute positional encodings for both image and text encoders. In contrast, TULIP adopts rotary positional encoding (RoPE)~\cite{su2024roformer} in its text encoder to support longer contexts. To examine how positional encoding schemes affect positional bias, we conduct experiments with TULIP on Urban1K and DOCCI. Results for text-based bias appear in Figure~\ref{fig:bias-text}(c,f), and image-based results are shown in Figure~\ref{fig:bias-image}(c,f). Compared to Long-CLIP, which uses absolute encoding, TULIP exhibits a stronger positional bias. For example, on Urban1K, shifting the first segment from position 0 to 2 causes TULIP’s retrieval accuracy to drop sharply from 0.66 to 0.065, while Long-CLIP shows a more modest decline from 0.582 to 0.485. \textit{These results highlight that the choice of positional encoding significantly influences positional bias: although absolute encodings mitigate the effect, neither strategy eliminates it entirely.} Interestingly, although both the text and image encoders in Long-CLIP use the same absolute positional encoding, the text encoder shows a clear bias toward the beginning, while the image encoder exhibits bias toward both the beginning and end.

\vspace{0.3ex}
\noindent{\bf Text Length.}
While our earlier experiments focus on multimodal models trained with long captions, it remains unclear whether similar positional bias arises when training uses only short-caption image-text pairs. To explore this, we evaluate CLIP-ViT-B/16 on the COCO dataset, where captions average 11.53 valid tokens. Due to the short captions, we analyze the first 12 tokens—split into three segments—and shift them across five positions spanning the full context window. As shown in Figure~\ref{fig:bias-text}(h), CLIP-ViT-B/16 still shows positional bias in the text encoder, favoring early positions, though the effect is weaker. A similar pattern appears in the image encoder, with higher accuracy at both the beginning and end (Figure~\ref{fig:bias-image}(h)). \textit{These findings suggest that positional bias persists regardless of caption length.}

\vspace{0.3ex}
\noindent{\bf Model Size.} To further validate the prevalence of such bias, we repeat the experiments across models of different parameter sizes, including CLIP-ViT-B/16 and CLIP-ViT-L/14. From Figure~\ref{fig:bias-text}(i) and Figure~\ref{fig:bias-image}(i), \textit{it is obvious that such bias is prevailing irrespective of model size}. In fact, CLIP-ViT-L/14 exhibits even stronger positional bias in both the text and image encoders, as evidenced by higher coefficient of variation across different positions compared to CLIP-ViT-B/16 (see Table~\ref{tab:segment_results} in the supplementary material). Beyond overall bias strength, we also observe differences in how positional information is retained across models. In the image encoder bias analysis, CLIP-ViT-L/14 maintains high retrieval accuracy when a segment is shifted back to its original position—typically lower than at the beginning and end, but still higher than at other positions. This phenomenon is either absent or substantially less pronounced in CLIP-ViT-B/16, suggesting that CLIP-ViT-L/14 retains more positional information, which may contribute to its superior retrieval performance. Notably, a similar pattern can also be observed in the visual positional bias analysis of TULIP (Figure~\ref{fig:bias-image}(c,f)), which is also based on the ViT-L/14 architecture.

{\setlength{\abovecaptionskip}{5pt}
\begin{figure*}[t]
    \centering
    \includegraphics[width=0.98\textwidth]{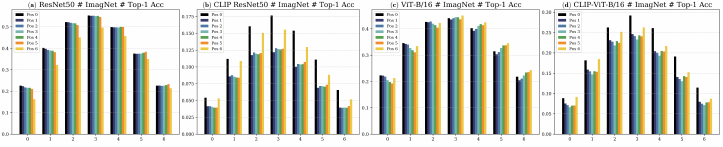}
    \caption{Classification performance on ImageNet for vision-only and CLIP-based models. We report Top-1 accuracy for ResNet-50 and ViT-B/16, both as standalone vision models and as vision backbones within CLIP.}
    \label{fig:classification}
    \vspace{-15pt}    
\end{figure*}

\noindent{\bf Image Resolution and Patch Size.}
Within the CLIP series, we further investigate the effects of image resolution and patch size on positional bias. A comparison between CLIP-ViT-B/16 and CLIP-ViT-B/32 (Figure~\ref{fig:bias-text}(g)) and Figure~\ref{fig:bias-image}(g)) reveals that positional bias remains evident and follows a similar distribution pattern. Notably, CLIP-ViT-B/32 shows a stronger bias in the image encoder, as reflected by the higher coefficient of variation reported in Table~\ref{tab:segment_results} in the supplementary material. Furthermore, comparing CLIP-ViT-L/14 with CLIP-ViT-L/14@336 (Figure~\ref{fig:bias-text}(j)) and Figure~\ref{fig:bias-image}(j)) suggests that increasing image resolution helps reduce bias for the image encoder. Although the highest accuracy consistently occurs when segments are moved to the beginning (positions 0 or 1) or end (position 6), indicating persistent bias, CLIP-ViT-L/14@336 exhibits a lower coefficient of variation across visual segment positions in 5 out of 7 cases (see Table~\ref{tab:segment_results}). In summary, reducing the patch size or increasing the image resolution is beneficial for lowering positional bias in the vision encoder.

\vspace{0.3ex}
\noindent{\bf Training Loss.} Previous studies have suggested that positional bias arises and worsens due to contrastive pretraining and fine-tuning \cite{coelho2024dwell}. In this paper, we examine the role of contrastive loss in multimodal models. Unlike CLIP and its variants, which use a softmax-based contrastive loss, SigLIP employs a sigmoid cross-entropy loss as its training objective \cite{zhai2023sigmoid}. After performing similar image-side and text-side bias analyses, we observe that positional bias still persists, as shown in Figure~\ref{fig:bias-text}(l) and Figure~\ref{fig:bias-image}(l). Interestingly, the bias pattern on the image side differs from that observed in CLIP-based models. SigLIP exhibits a prominent bias toward the beginning of the sequence. Although the accuracies at the final positions are generally higher than at the penultimate position, they are typically lower than those at most other positions. \textit{These findings suggest that the training loss plays an important role in positional bias, as contrastive learning and sigmoid loss lead to different bias patterns}.

\vspace{0.3ex}
\noindent{\bf Model Structure.}
Given the experiments above, a natural question arises: are these biases specific to transformer-based models? To address this, we conducted an additional experiment using CLIP-ResNet-50, where the vision encoder is a ResNet-50 — a convolutional neural network (CNN) rather than a transformer. As presented in Figure~\ref{fig:bias-text}(k) and Figure~\ref{fig:bias-image}(k), positional bias still persists, and the observed patterns resemble those seen with ViT-based image encoders. \textit{This suggests that positional bias is not exclusive to transformer-based models}. Future work could explore whether this holds when both image and text encoders are non-transformer-based.

\vspace{0.3ex}
\noindent{\bf Uni-Modality vs. Multi-Modality.} Coelho  \etal \cite{coelho2024dwell} reported a “dwelling at the beginning” bias in certain text representation models. Interestingly, this bias pattern in unimodal models trained solely on text data is consistent with the pattern we observe in the text encoder of multimodal models trained on image-text pairs. Here, we shift focus to the image side, investigating whether the existence and pattern of positional bias remain consistent when the vision encoder is trained with image-only data versus image-text pairs. To this end, we adopt image classification on ImageNet \cite{deng2009imagenet} as the evaluation task and utilize Top-1 accuracy as the metric. For multimodal models, we use zero-shot classification, where class categories are formatted as natural language prompts. As shown in Figure~\ref{fig:classification}(a), the accuracy of moving a segment across different positions using ResNet classification model remains relatively stable, indicating no clear positional bias. This finding is expected as CNN models usually possesses the property of translation invariance. In contrast, CLIP-ResNet-50 zero-shot classification results in Figure~\ref{fig:classification}(b) exhibit clear positional bias at the beginning and end. To further validate this observation, we conducted the same comparison using ViT-B/16 as the vision encoder and found that the positional bias is significantly more pronounced under the CLIP framework (see the comparison between Figure~\ref{fig:classification}(c) and Figure~\ref{fig:classification}(d)). \textit{These findings suggest that the positional bias on the image side in multimodal models originates from, or is amplified by, training with image–text pairs}.

\section{Conclusion and Future Work}
This paper presents an empirical analysis of positional bias in multimodal representation models, mainly focusing on image-text retrieval. We find that positional bias is widespread: text encoders tend to emphasize early positions, while image encoders exhibit bias at the beginning or both ends. This bias appears to arise from—or be shaped by—multiple factors, including positional encoding schemes, the contextual importance of training data, training objectives, and the intrinsic use of image-text pairs in multimodal learning.

For future work, we aim to further investigate the underlying causes of positional bias and develop strategies to mitigate it. For instance, the Differential Transformer \cite{ye2024differential}, which introduces a differential attention mechanism to reduce attention noise in causal models, may offer promising directions if adapted to bidirectional multimodal representations.

\bibliography{aaai2026}

%


\clearpage
\appendix
\section*{Supplementary Material}

\setcounter{section}{0} 
\renewcommand{\thesubsection}{\thesection.\arabic{subsection}} 

\section {Related Works}
\noindent{\bf Positional Bias in Text Generation.}
Most studies on positional bias have focused on text generation models. Several works have explored different aspects of this issue, including the identification of the phenomenon, analysis of its underlying causes, and investigation of potential mitigation strategies.
In \cite{liu2024lost}, the 'lost in the middle' phenomenon was identified for the first time by analyzing performance of long-context language models on tasks of multi-document question answering and key-value retrieval. This finding also inspired the authors to propose new model evaluation protocols.  In \cite{hsieh2024found}, it is found that LLMs usually exhibit an U-shaped attention bias, with the tokens at the beginning and at the end given higher attention, irrespective of their contextual importance. To alleviate such bias, a calibration mechanism called found-in-the-middle was proposed so that the allocated attention can be faithful to the actual relevance. Ye  \etal \cite{ye2024differential} noticed that Transformer usually allocates a small portion of attention to relevant regions, while putting a majority of its attention on irrelevant context. They proposed to handle this challenge by  introducing a modified architecture, namely DIFF Transformer, which adopts a differential attention mechanism to cancel attention allocated to irrelevant context, but amplify these for the relevant context. In \cite{wang2024eliminating}, it was claimed that position bias is the results of two common modules in LMs: causal attention and position embedding. The resulted solution was to replace the causal attention with bidirectional attention, and also re-order the documents based on model attention values.  In \cite{he2023never}, the approach to overcoming the bias is augmenting the training documents so that the correct answers are located at arbitrary positions in contexts among noisy documents. In \cite{peysakhovich2023attention}, it was discovered that most models show bias towards context that are close in position to the generated response. And they put forward to sort documents based on attention scores before running model generation. 

Most existing studies on positional bias have concentrated on text generation models, investigating different facets of this issue, including its identification, underlying causes, and possible mitigation strategies. The ``lost in the middle'' phenomenon, first identified by Liu  \etal~\cite{liu2024lost}, highlights how long-context language models often underperform when relevant information appears in the middle of an input sequence. Their analysis, conducted on multi-document question answering and key-value retrieval tasks, also motivated the development of new evaluation protocols for long-context understanding. Hsieh  \etal~\cite{hsieh2024found} further revealed that large language models (LLMs) tend to exhibit a U-shaped attention pattern, prioritizing tokens at the beginning and end of sequences regardless of their contextual relevance. To address this, they proposed a calibration mechanism called \textit{found-in-the-middle}, which adjusts attention to more accurately reflect the true importance of tokens.

In a related line of work, Ye  \etal~\cite{ye2024differential} observed that standard Transformers often allocate disproportionate attention to irrelevant context, while neglecting relevant regions. They introduced the DIFF Transformer, a modified architecture that employs a differential attention mechanism to suppress attention on unimportant tokens while amplifying it for relevant ones. Similarly, Wang  \etal~\cite{wang2024eliminating} attributed positional bias to two architectural components: causal attention and position embeddings, and proposed a solution that replaces causal attention with bidirectional attention, combined with document reordering based on attention distributions. He  \etal~\cite{he2023never} approached the problem from a data-centric perspective by augmenting training data such that correct answers appear at arbitrary positions within noisy contexts, encouraging models to attend to content irrespective of position. Similarly, Peysakhovich  \etal~\cite{peysakhovich2023attention} recommended sorting documents by attention relevance prior to response generation to reduce this bias.

Tian  \etal~\cite{tian2025identifying} extend the bias analysis from LLMs to large vision-language models (LVLMs), with a particular emphasis on models capable of reasoning over multiple images.  Their findings reveal a positional bias favoring image placement toward the end of the input sequence in open-source models, which they attribute to the use of causal attention. While the study is conducted in the multimodal domain, the underlying architecture remains aligned with text generation models, treating visual inputs as special text tokens and focusing primarily on question answering tasks, which are inherently text generation-based.

Prior work has consistently demonstrated the pervasive nature of positional bias in text generation models, offering a range of perspectives, including architectural modifications, training strategies, and inference-time techniques, to understand and mitigate its effects. In contrast, our work investigates positional bias in multimodal representation learning, where we observe distinct bias patterns that differ notably from those found in text generation models.

\begin{figure*}[htb]
    \centering
    \includegraphics[width=0.98\textwidth]{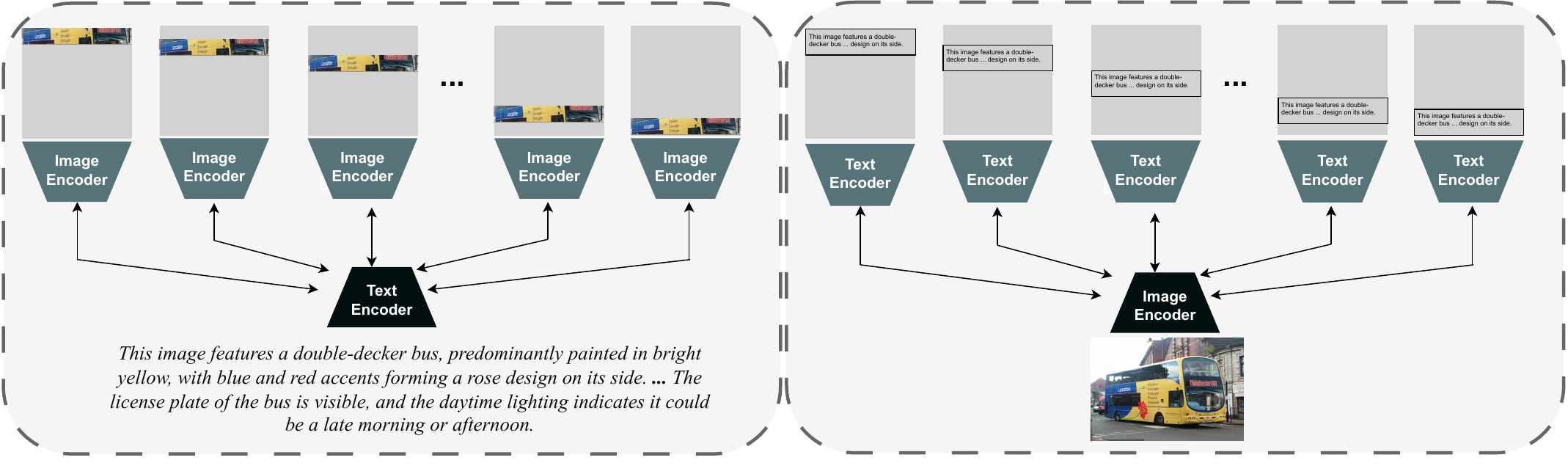}
    \caption{Demonstration of the experimental setup for analyzing positional bias in image and text encoders. For image bias (left), a fixed visual segment is shifted across spatial positions while the caption remains constant. For text bias (right), a fixed text segment is moved across positions in the input sequence with the image held constant. Retrieval scores are computed at each position in both cases. }
    \label{fig:image-to-text}
\end{figure*}

\begin{figure*}[tb]
    \centering
    \includegraphics[width=0.98\textwidth]{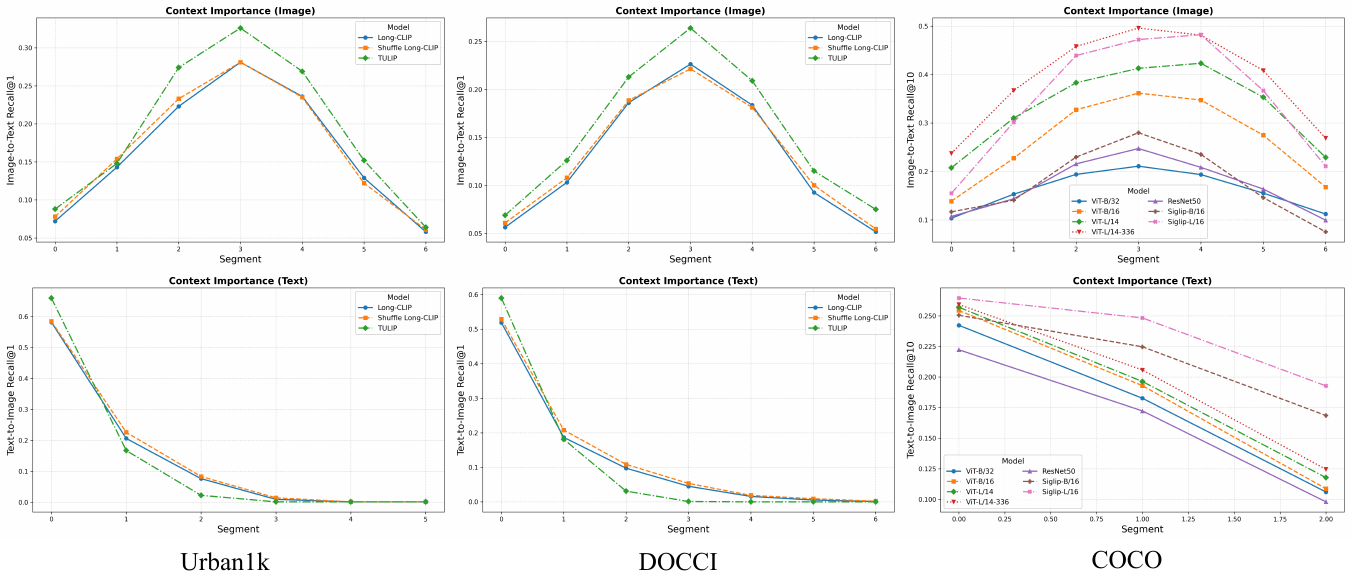}
    \caption{Context importance of different modalities in retrieval tasks. }
    \label{fig:context_importance_more}
    \vspace{-5pt}
\end{figure*}

{\setlength{\abovecaptionskip}{10pt}
\begin{figure*}[tb] 
    \centering
    \includegraphics[width=0.98\textwidth]{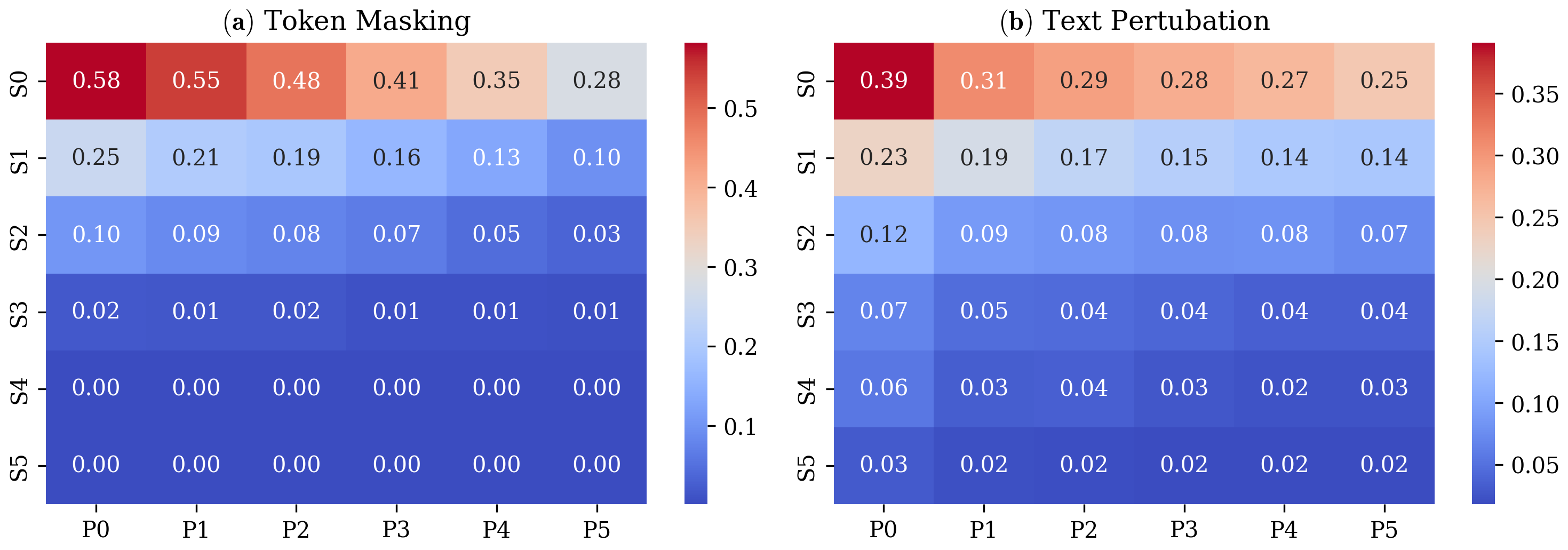}
    \caption{Comparison of text-to-image retrieval using token masking (left) vs. text perturbation (right). Each row shows a different segment shifted across positions; different rows correspond to different segments.}
    \label{fig:dummy-token-comparsion} 
     \vspace{-5pt}
\end{figure*}

\noindent{\bf Positional Bias in Text Representation.}
Coelho  \etal \cite{coelho2024dwell} investigated positional biases in text representation modeling, focusing on web document retrieval tasks. Unlike text generation models, their study revealed that text representation models tend to emphasize the beginning. Interestingly, this bias was not found to be a result of language modeling pretraining but emerged after contrastive pretraining. The phenomenon, referred to as "dwell in the beginning," was attributed to the inverted pyramid writing style, where the most important information is typically presented at the start of a document. In \cite{ben2024impact}, the positional bias of language representation models in token classification tasks, such as Named Entity Recognition (NER) and Part-of-Speech (POS) tagging, is investigated. A similar tendency was observed, where models exhibit a bias toward the beginning of sequences. The authors hypothesize that this bias arises due to properties of the training data: the sequences are generally short and the most informative tokens often appear early. Focusing on the application of extractive question answering with text embedding models, the study in \cite{ko2020look} also reveals a positional bias toward the beginning of the input. The authors attribute this bias to the characteristics of the training data distribution. The study in \cite{goel2024quantifying} further demonstrates the presence of positional bias in text embedding models, specifically from the perspective of semantic similarity measurement. Across eight different models, regardless of their positional encoding mechanisms, it was consistently observed that the initial portions of input texts are disproportionately emphasized. This bias was attributed to input truncation during preprocessing, a common practice used to constrain inputs within the model’s context window. In this paper, we extend the investigation of positional bias to the multimodal setting by empirically examining its existence, patterns, and the underlying causes or contributing factors.

\begin{figure*}[tb]
    \centering
    \includegraphics[width=0.98\textwidth]{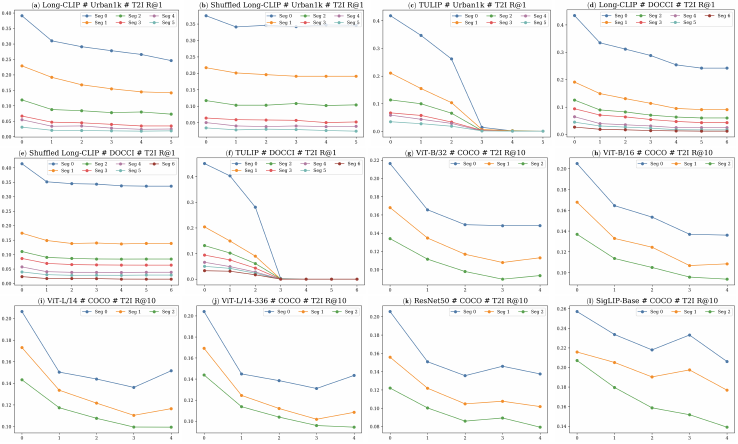}
    \caption{Positional bias analysis of text encoders across multimodal models and datasets using text perturbation strategy. Each figure title follows the format: model name \# dataset \# metric. The x-axis shows positions, and colors indicate different segments.}
    \label{fig:bias-text-perturbation}
    \vspace{-15pt}     
\end{figure*}

\begin{table*}[tb]
\centering
\begin{tabular}{|p{4.4cm}|p{1.3cm}|p{1.3cm}|p{1.3cm}|p{1.3cm}|p{1.3cm}|p{1.3cm}|p{1.3cm}|}
\hline
\textbf{Model} & \textbf{seg0} & \textbf{seg1} & \textbf{seg2} & \textbf{seg3} & \textbf{seg4} & \textbf{seg5} & \textbf{seg6} \\
\hline
Long-CLIP\textsuperscript{\dag}          & 0.146 & 0.040 & 0.084 & 0.087 & 0.099 & 0.098 & 0.159 \\
\hline
Shuffled Long-CLIP\textsuperscript{\dag} & 0.185 & 0.054 & 0.077 & 0.089 & 0.098 & 0.106 & 0.144 \\
\hline
TULIP\textsuperscript{\dag}              & 0.220 & 0.098 & 0.098 & 0.107 & 0.179 & 0.112 & 0.186 \\
\hline
Long-CLIP\textsuperscript{\ddag}           & 0.140 & 0.125 & 0.095 & 0.110 & 0.158 & 0.199 & 0.223 \\
\hline
Shuffled Long-CLIP\textsuperscript{\ddag}   & 0.155 & 0.148 & 0.102 & 0.116 & 0.157 & 0.190 & 0.226 \\
\hline
TULIP\textsuperscript{\ddag}                & 0.308 & 0.184 & 0.161 & 0.188 & 0.316 & 0.231 & 0.170 \\
\hline
ViT-B/32              & 0.163 & 0.158 & 0.175 & 0.178 & 0.212 & 0.243 & 0.274 \\
\hline
ViT-B/16              & 0.060 & 0.084 & 0.096 & 0.094 & 0.100 & 0.116 & 0.136 \\
\hline
ViT-L/14              & 0.162 & 0.118 & 0.107 & 0.104 & 0.160 & 0.133 & 0.121 \\
\hline
ViT-L/14--336        & 0.142 & 0.125 & 0.108 & 0.101 & 0.104 & 0.134 & 0.116 \\
\hline
ResNet-50          & 0.152 & 0.100 & 0.110 & 0.117 & 0.131 & 0.157 & 0.211 \\
\hline
SigLIP-Base                  & 0.289 & 0.126 & 0.099 & 0.109 & 0.118 & 0.148 & 0.232 \\
\hline
\end{tabular}%
\caption{Segment-wise variance coefficient across models. We report results for three models (Long-CLIP, Shuffled Long-CLIP, and TULIP) evaluated on two long-caption datasets (Urban1k\textsuperscript{\dag} and DOCCI\textsuperscript{\ddag}), as well as six short-caption models evaluated on COCO. }
\vspace{-15pt} 
\label{tab:segment_results}
\end{table*}

\noindent{\bf Multimodal Representation Learning.}
Unlike language models such as BERT \cite{devlin2019bert} and LLaMA \cite{touvron2023llama} that are limited to processing textual input, multimodal models are capable of handling inputs from multiple modalities by projecting them into a shared embedding space, thereby enabling the capture of cross-modal relationships. A prominent and pioneering example is CLIP \cite{radford2021learning}, which learns to align image and text representations through contrastive learning. Trained on 400 million image-text pairs, CLIP demonstrates strong performance across various downstream tasks, including zero-shot image classification and image-text retrieval. Despite its success, CLIP also has notable limitations. One major challenge lies in its reliance on contrastive loss, which necessitates large batch sizes to provide sufficient negative samples during training—resulting in significant computational overhead. To mitigate this issue, SigLIP \cite{zhai2023sigmoid} was proposed, introducing a sigmoid-based binary classification loss that eliminates the need for large batch sizes. Experimental results show that SigLIP achieves comparable or superior performance to CLIP on a range of vision-language tasks while simplifying the training process.

Despite the notable success of CLIP and its successor SigLIP, both models are trained predominantly on image-text pairs with relatively short textual descriptions. In practice, most CLIP variants constrain the input text length to 77 tokens. More notably, recent work has shown that only a small subset of these tokens are effectively utilized during training, with the number of well-trained tokens being fewer than 20 \cite{zhang2024long}. To address this limitation, Long-CLIP \cite{zhang2024long} introduces a knowledge-preserving positional embedding stretching strategy, extending the token limit to 248 and fine-tuning the model on a dataset specifically curated for long image-text pairs \cite{chen2024sharegpt4v}. In contrast to such approaches that rely on absolute positional encoding, TULIP \cite{najdenkoska2024tulip} adopts relative positional encoding, enabling the model to process text of arbitrary length. This design allows CLIP-like models to make full use of longer captions, thereby enhancing their ability to capture fine-grained and detailed semantic information embedded in extended textual contexts.

Other multimodal models, such as LLaVA \cite{liu2023visual} and BLIP-2 \cite{li2023blip}, have gained prominence for enabling complex, instruction-based multimodal generation tasks like visual question answering and image captioning. However, in this work, we focus on CLIP and its variants, which are designed for multimodal representation learning through contrastive alignment of image and text embeddings.  This focus is motivated by the fact that CLIP-style models frequently serve as the key backbone for more advanced multimodal systems, including those used in generative settings. In addition, cross-modal retrieval plays a crucial role in multimodal generation pipelines that incorporate Retrieval-Augmented Generation (RAG) techniques \cite{wu2024visual, xia2024mmed}, where retrieval mechanisms are employed to select relevant visual or textual content as input to the generation module. As such, a deeper understanding of CLIP-like models in the context of cross-modal retrieval is critical for advancing the broader landscape of vision-language learning. 

\section {Flowchart of Positional Bias Analysis}

Figure~\ref{fig:image-to-text} illustrates our approach for analyzing positional bias in multimodal models across both image and text modalities. For clarity, we show one example each for an image segment and a text split. In the image-side analysis (left), a single visual segment is isolated and shifted across spatial positions, with the caption fixed and all other image regions masked; retrieval scores are computed at each position. On the text side (right), a text split is selected and moved across different positions in the input text, while the image remains fixed and the rest of the text are masked. This process is repeated for all segments.

\section{More Experiments}

\subsection {Additional Experiments on Context Importance}\label{sec:supp_context_importance}

We present additional context importance experiments across image and text modalities (Figure~\ref{fig:context_importance_more}). For long-caption models, we include Long-CLIP, Shuffled Long-CLIP, and TULIP; for short-caption models, we evaluate CLIP with various backbones (ViT-B/32, ViT-B/16, ViT-L/14, ViT-L/14-336, ResNet-50) and SigLIP-Base. Experiments are conducted on the DOCCI and Urban1K datasets for long captions, and on COCO for short captions.

Across all settings, context importance trends remain consistent across datasets and architectures. For images, the central region is most important, with declining relevance toward the edges. For text, the first segment yields the highest retrieval accuracy, decreasing steadily in later segments.

\subsection {Token Masking vs. Text Perturbation}

We explore two strategies for analyzing positional bias in the text encoder: token masking and text perturbation. Figure~\ref{fig:dummy-token-comparsion} compares text-to-image retrieval performance under both methods on Urban1K using Long-CLIP. Both approaches yield similar trends across positions (x-axis), consistently revealing strong positional bias toward the beginning.

Furthermore, Figure~\ref{fig:bias-text-perturbation} shows positional bias trends across different multimodal models and datasets using the text perturbation strategy, revealing patterns consistent with those observed in the token masking experiments.

\subsection{Measurement of Positional Bias}
To quantitatively assess positional bias, we use the coefficient of variation to capture how retrieval accuracy changes when the same segment is shifted across positions, with all other positions masked. Table~\ref{tab:segment_results} reports this metric for various models and datasets in our image-side analysis. While the values offer useful insights, our primary goal is to highlight the presence and patterns of positional bias. A more precise quantification and mitigation of this bias is left for future work.

\end{document}